\documentclass[10pt,twocolumn,letterpaper]{article}

\usepackage[pagenumbers]{cvpr} 

\usepackage{graphicx}
\usepackage{amsmath}
\usepackage{amssymb}
\usepackage{booktabs}
\usepackage{dsfont}
\usepackage{amsmath}
\usepackage{enumitem}
\usepackage{amssymb}
\usepackage{tabulary,overpic}
\usepackage{multicol}
\usepackage{multirow}

\usepackage[pagebackref,breaklinks,colorlinks]{hyperref}

\usepackage[capitalize]{cleveref}
\crefname{section}{Sec.}{Secs.}
\Crefname{section}{Section}{Sections}
\Crefname{table}{Table}{Tables}
\crefname{table}{Tab.}{Tabs.}

\def\confName{CVPR}
\def\confYear{2022}

\def\CNA{{CNA}}
\DeclareMathOperator*{\argmax}{arg\,max}
\DeclareMathOperator*{\argmin}{arg\,min}

\newcommand{\Acc}{{Acc}}

\newlength\savewidth

\begin{document}

\title{Contrastive Neighborhood Alignment}

\author{Pengkai Zhu\thanks{equal contribution}\\
{\it Amazon/AWS AI}
\and
Zhaowei Cai\footnotemark[1]\\
{\it Amazon/AWS AI}
\and 
Yuanjun Xiong\footnotemark[1]\\
{\it Amazon/AWS AI}
\and
Zhuowen Tu\footnotemark[1]\\
{\it Amazon/AWS AI}
\and
Luis Goncalves\\
{\it Amazon/AWS AI}
\and
Vijay Mahadevan\\
{\it Amazon/AWS AI}
\and
Stefano Soatto\\
{\it Amazon/AWS AI}
}
\maketitle

\begin{abstract}
    We present  Contrastive Neighborhood Alignment (CNA), a manifold learning approach to maintain the topology of learned features whereby data points that are mapped to nearby representations by the source (teacher) model are also mapped to neighbors by the target (student) model. The target model aims to mimic the local structure of the source's representation space using a contrastive loss. CNA is an unsupervised learning algorithm that does not require ground-truth labels for the individual samples. CNA is illustrated in three scenarios: manifold learning, where the model maintains the local topology of the original data in a dimension-reduced space; model distillation, where a small student model is trained to mimic a larger teacher; and legacy model update, where an older model is replaced by a more powerful one. Experiments show that CNA is able to capture the manifold in a high-dimensional space and improves performance compared to the competing methods in their domains. 
\end{abstract}

\section{Introduction}
\begin{figure}
    \centering
    \includegraphics[width=0.85\linewidth]{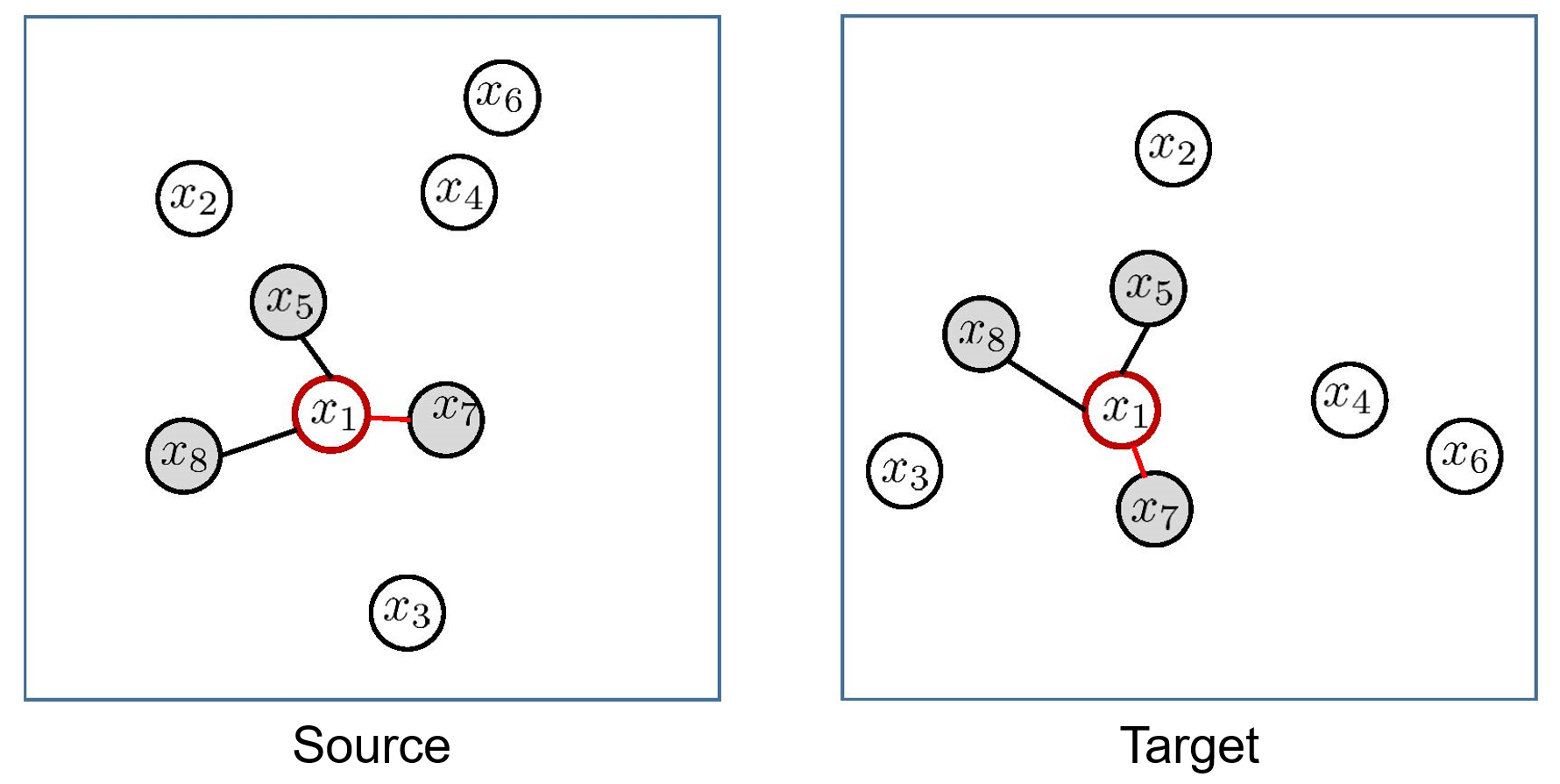}
    \caption{The local topology of feature space in source (left) and target (right) models. \CNA{} preserves the local structure between models, such that the neighbors ($x_5, x_7, x_8$ for $x_1$) from the source model stay the same in the target model.}
    \label{fig:CNA_illustration}
    \vspace{-5mm}
\end{figure}

In this paper, we present a new algorithm, Contrastive Neighborhood Alignment (\CNA), that preserves the local topology of learned features between source and target models by mapping neighbors in one space (source) to neighbors also in the target (student) space. \CNA{} overcomes the optimization challenge in the formulation of traditional manifold learning methods for large-scale computing by designing a contrastive loss that can be trained effectively to preserve the local topology of the feature space. The model learned through \CNA{} is inductive and hence can generalize to novel data. \CNA{} is an unsupervised approach and requires no ground-truth labels. \cref{fig:CNA_illustration} illustrates the local topology preservation by \CNA, that the neighbors ($x_5, x_7, x_8$) of a sample $x_1$ (\cref{fig:CNA_illustration} left) in the source space is well-preserved in the target model space trained by \CNA (\cref{fig:CNA_illustration} right).

\CNA{} does not impose constraints on source models and it can be adopted to performing student-teacher learning for tasks such as knowledge distillation (KD) \cite{hinton2015distilling} and reducing model regression in model updating \cite{yan2020positive}, beyond the standard manifold learning tasks \cite{roweis2000nonlinear,donoho2003hessian,tenenbaum2000global}. Conventional knowledge distillation and student-teacher (S-T) learning \cite{hinton2015distilling,wang2021knowledge,dkozlovgithub,cheng2017survey} approaches primarily follow the design principle of matching features or classification distributions between the teacher and student models \cite{wang2021knowledge}. Despite the effectiveness of the feature/distribution matching criterion, S-T learning methods are not without limitations. Typically, student-teacher training tries to force the student to mimic the input-output behavior of the teacher, but the regularization term (model bias) often results in a direct conflict with the cross-entropy (CE) loss, i.e., KL-divergence/logit-matching vs. softmax. This has two problems.
First, the criterion trades off error rate minimization. Second, the criterion is too rigid in that it mimics the teacher's predictions without the awareness of the data manifold. In contrast, \CNA{} tackles the student-teacher learning problem by modeling the knowledge in the models using the local structures in their feature space, which is distinct from the existing methods in model distillation \cite{hinton2015distilling}.

\CNA{} is a new inductive principle: instead of mimicking the input-output behavior, it transfers the local neighborhood structure. Neighbors remain neighbors, when transferring from the source to the target, regardless of the label. This has two distinct advantages: first, it does not require ground truth labels. Second, it allows more flexibility to deviate from the inference of the source model when it is mistaken. In \CNA{}, we impose that samples are neighbors in the source (teacher) representation space remain neighbors in the target (student) representation space. This alignment of local neighborhoods is achieved by adding to the cross-entropy loss a contrastive term that is in the same form (softmax function) as the cross-entropy term, making the optimization well conditioned; as both the cross-entropy and the contrastive loss are in similar unit spaces (normalized softmax function), balancing the two terms becomes relatively easy.
The contributions of our work are listed below:
{
\begin{itemize}
 \setlength\itemsep{0mm}
 \setlength{\itemindent}{0mm}
    \item We propose a new method for manifold learning that maintains the local neighborhood in inductive models.
    \item We introduce an instance-level contrastive loss for preserving the neighborhood structures that is different from how contrastive loss is motivated and implemented previously \cite{wu2018unsupervised,oord2018representation,he2016deep}.
    \item We propose to use a new loss function in Student-Teacher training, which is distinct from the existing methods in this field.
\end{itemize}

The effectiveness of \CNA{} is demonstrated on three tasks, 1) dimensionality reduction for manifold learning \cite{van2009dimensionality}, 2) model distillation \cite{hinton2015distilling}, and 3) model update regression minimization \cite{yan2020positive}. \CNA{} is capable of capturing the manifold from the source space and can be applied to various downstream tasks in knowledge distillation.}

\section{Related work}
\label{sec:related}

\noindent {\bf Manifold/metric learning}: Preserving local neighborhood structure is a popular direction in machine learning that has been explored in both metric learning \cite{yang2006distance,davis2007information,kulis2012metric,hoffer2015deep} and manifold learning \cite{roweis2000nonlinear}. Standard manifold learning approaches perform dimensionality reduction by mapping the feature representation from the original space to the new one \cite{van2009dimensionality}. They attempt to maintain the local structure through the reconstruction by the convex combination of same neighbors \cite{roweis2000nonlinear,donoho2003hessian}, by retaining the neighborhood distances while unfolding the manifold \cite{weinberger2004learning}, or by preserving pairwise geodesic distances between all data points \cite{tenenbaum2000global}. These manifold learning methods learn the manifold transductively, which limits their deployments in modern deep networks. In contrast, Our \CNA{} adopts an instance-level contrastive loss that attempts to preserve the manifold structure of feature space in an inductive model, which is significantly different from these works. Additionally, \CNA{} can be combined with classification loss (e.g. a cross-entropy loss) to create classifiers when training deep models.

\noindent {\bf Contrastive learning and self-supervised learning}: Contrastive learning~\cite{hadsell2006contrastive} is used for either supervised~\cite{taigman2014deepface, schroff2015facenet,khosla2020supervised} or unsupervised~\cite{wu2018unsupervised,chen2020simple,he2020momentum} representation learning. It has recently become popular in self-supervised learning through the instance discrimination pretext task~\cite{wu2018unsupervised}. Contrastive representation distillation (CRD)~\cite{tian2020contrastiverep} extends the idea of contrastive learning to the teacher-student training case.
Though not directly minimizing a certain distance function, CRD still needs to calculate the distance between student's feature vectors and that of the teacher.
\CNA{} uses the contrastive loss form in teacher student training but does not measure the ``cross model distance''. It only concerns the local structures within the models' feature spaces. The supervised contrastive learning method \cite{khosla2020supervised} is also different from \CNA{} since we focus on preserving the local neighborhoods between models as opposed to using class labels to learn visual representation.

\noindent {\bf Model distillation}: There has been a significant amount of work to study how knowledge can be shared between multiple models. Model distillation~\cite{hinton2015distilling,wang2021knowledge} transfers knowledge from a larger, more powerful teacher model to a smaller student model by matching their output/representation. 
\CNA{} does not rely on matching outputs~\cite{hinton2015distilling,yan2020positive} or reusing the old models’ parameters~\cite{Shen_2020_CVPR}. Instead, knowledge is shared between models through the information encoded in the neighborhood structure of a model's feature space.

\noindent {\bf Regression minimization for model upgrading}: The presence of new errors that were not manifest when using the old model (negative flips) can cause a perceived reduction of performance by users, referred to in the industry as ``regression.'' 
Although progress in the design of the architecture of backbone models \cite{szegedy2015going,simonyan2015very,he2016deep,xie2017aggregated,howard2017mobilenets,hu2018squeeze,vaswani2017attention}, new optimization and regularization schemes \cite{krizhevsky2012imagenet,duchi2011adaptive,ba2016layer}, and the availability of new datasets \cite{DBLP:journals/ijcv/RussakovskyDSKS15,kuehne2011hmdb,lin2014microsoft,antol2015vqa,rajpurkar2016squad,krishna2017visual} have driven a fast-paced reduction of the average error rate (AER) of image classifiers,
regression is still a major obstacle to the deployment of improved models. Even a modest number of negative flips can nullify the benefit of a large decrease in AER. 
Overall, the problem of reducing the negative flip rate (NFR), along with the average error rate (AER), has been referred to as Positive-Congruent Training, or PC-Training~\cite{yan2020positive}. Minimizing model regression during model upgrading is an emerging problem that is of great practical importance~\cite{yan2020positive}. In~\cite{yan2020positive}, NFR is proposed as the metric to measure regression in classification model updates and an approach based on focal distillation, a variant of model distillation~\cite{hinton2015distilling}, is proposed to reduce regression. Our work addresses the same problem through contrastive neighbor alignment instead of relying on model distillation.

\section{Contrastive Neighborhood Alignment}
\label{sec:method}

\begin{figure*}
    \centering
    \includegraphics[width=0.82\textwidth]{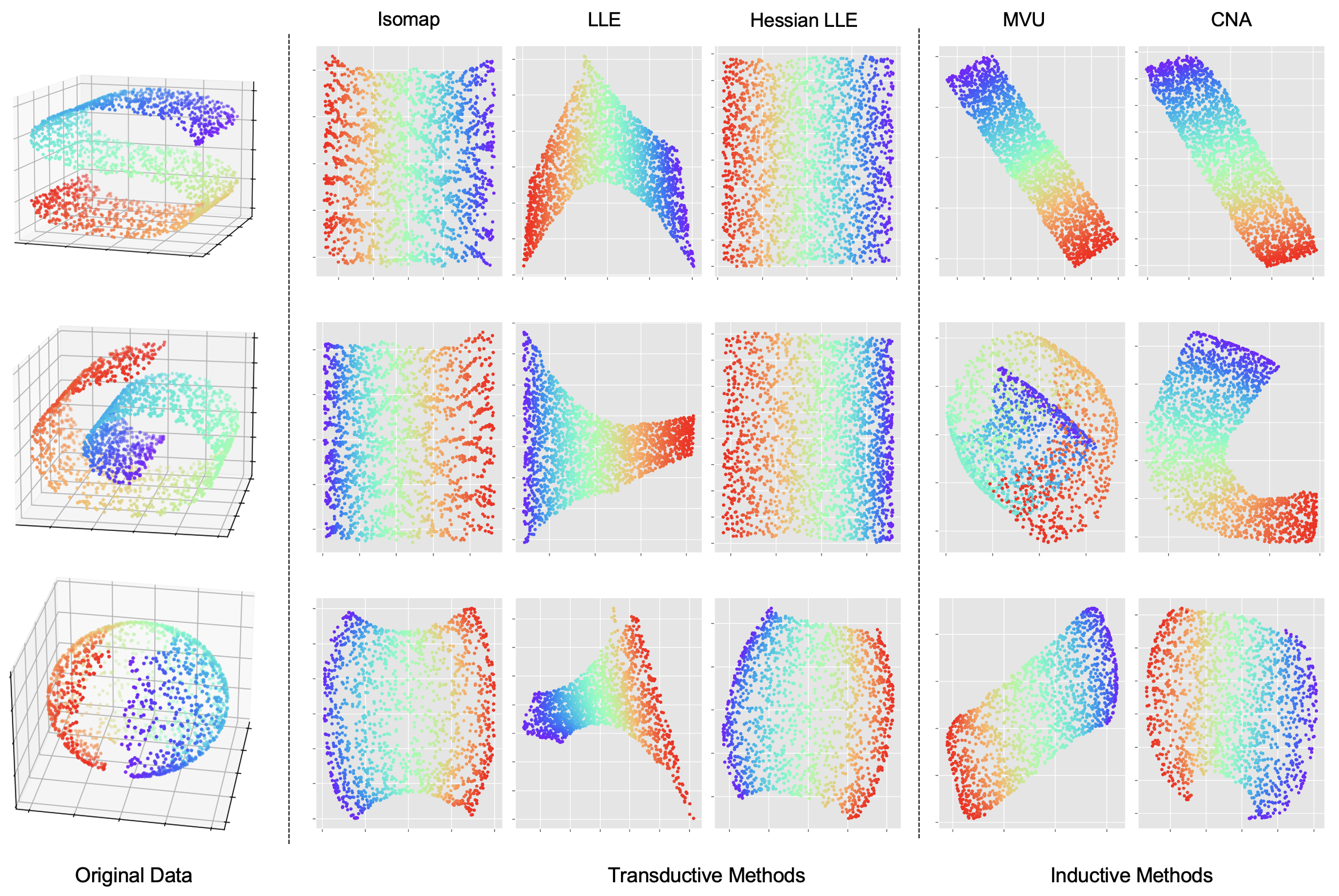}
    \vspace{-4mm}
    \caption{Qualitative results of manifold learning on synthetic datasets. Rows(top to bottom): S-curve, Swiss Roll, Sphere. Columns(left to right): original 3D data points, transductive methods (Isomap \cite{tenenbaum2000global}, LLE \cite{roweis2000nonlinear}, Hessian LLE \cite{donoho2003hessian}), inductive methods (MVU \cite{weinberger2004learning}, \CNA{}).}
    \label{fig:manifold}
    \vspace{-4mm}
\end{figure*}

Given a possibly pretrained source model $S$ and a dataset $\mathcal{D} = \{x_i\}$, it is a common task to transfer information from the source model $S$ to a newly trained target model $T$. Let $f_S(\cdot)$ and $f_T(\cdot)$ denote the feature representation of the source and target model, respectively. 
The two models could have different architectures or different feature dimensions. For example, in manifold learning \cite{van2009dimensionality}, the high-dimensional source feature space is cast to a target space in which the low-dim manifold is unveiled. In model distillation, knowledge is shared between models by minimizing the sample-wise pseudo-distance between the two models' classification posterior probabilities. In this work, we propose \CNA{}, a contrastive loss that transfers information between models by preserving of the local structures of the source model feature space in that of the target. An illustration of this idea is shown in \cref{fig:CNA_illustration}. Below, we detail the derivation of the \CNA{} loss function and its applications in model distillation/update tasks.

\subsection{Preserving Local Structure}
\label{sec:LLE}

The idea of replicating the local structure in a new feature space has been explored in the manifold learning literature \cite{van2009dimensionality}. For example, Isomap \cite{tenenbaum2000global} preserves pairwise geodesic (or curvilinear) distances between samples. The geodesic distances are computed by constructing a neighborhood graph in which every sample $x_i$ is connected with its top-$K$ nearest neighbors. Locally linear embedding (LLE) \cite{roweis2000nonlinear} studied the local structure in the non-linear feature space. It has shown that by embedding high-dimensional feature vectors in their local neighborhoods, we can obtain effective low-dimensional embedding of the data points. LLE assumes each data point and its neighbors lie on a locally linear patch in the source space, and characterizes the local structure by a linear reconstruction using the neighbors:
\begin{equation}
    \epsilon(W) = \sum_i \| f_S(x_i) - \sum_{j}W_{i,j}f_S(x_j) \|_2^2 \label{eq:lle_project}
\end{equation}
where $f_S(x_i)$ is the feature representation of $x_i$ in the source space, 
$\epsilon$ is the reconstruction error and $W$ is the coefficient matrix. $W_{i,j}$ is enforced to 0 if $x_j$ is not one of the top-$K$ neighbors of $x_i$ and $\sum_j W_{i,j}=1$. The solution $W^* = \argmin \epsilon(W)$ is then used to obtain the target representation by minimizing the reconstruction error in the target space. Maximum Variance Unfolding (MVU) \cite{weinberger2004learning} is another work that focuses on retaining pairwise distances in a neighborhood graph. It maximizes the sum of distances between all data points in the target space:
\begin{equation}
    \max \sum_{i,j} d(f_T(x_i),f_T(x_j)), \\
\end{equation}
under the constraint that the distances between the neighbors are preserved: $d(f_T^{(i)}, f_T^{(j)}) = d(f_S^{(i)}, f_S^{(j)})$ if $x_j$ is one of the top-$K$ neighbors of $x_i$. Here $d(\cdot, \cdot)$ is the $\ell_2$ distance. 

However, these manifold learning methods are difficult to be employed in modern deep learning frameworks. They often need to store a local structure graph for the entire dataset, which occupies a lot of memory. Inferring the representation for each data point is also not viable when the dataset is large. The features are assumed to live in the same metric space defined by the $\ell_2$ distance function, which is often not the case for highly non-linear deep neural networks. The generalization to novel data is also difficult due to the transductive nature of these methods (see discussions in \cref{sec:cna_loss}). 

Inspired by these approaches, and to remedy these issues, \CNA{} is proposed to preserve the local structure in the target model's feature space by enforcing the same neighbors from the source space.

\subsection{Contrastive Neighborhood Alignment Loss}
\label{sec:cna_loss}

We hypothesize that the knowledge can be shared between two models when the nearest neighbors of $x_i$ in the source model's feature space are still its nearest neighbors in the target model's feature space. For simplicity, we can set $K=1$ in \cref{eq:lle_project}, resulting in each sample's feature $f_T(x_i)$ being represented only by its nearest neighbor $f_T(x_{i^\star})$ in the feature space, with $W_{ii^\star}=1$. Then the learning objective can make the target model maintain the local structure encoded by $W$ in its own feature space. By resorting to instance-based contrastive loss, we propose the contrastive neighborhood alignment (\CNA) loss that achieves the goal of matching the same neighbors in an easy to optimize manner.

Specifically, for each sample $x_i$ in a batch of $M$ samples, we locate its nearest neighbor $\aleph(x_i)$ in the feature space of the source model:
\begin{equation}
    \aleph(x_i) = x_{i^\star} \label{eq:nb_u},
\end{equation}
where
\begin{equation}
    i^\star = \argmax_{j=1,\ldots,M, j\ne i} \frac{f_S(x_i)\cdot f_S(x_j)}{|f_S(x_i)|_2\cdot |f_S(x_j)|_2}, \label{eq:nb_index}
\end{equation}
where $f_S(\cdot)$ is the feature representation of the source model and $|\cdot|_2$ is the $\ell_2$ norm of the feature. The nearest neighborhood is the indicator we use for the model local topology. And the contrastive neighborhood alignment loss is designed to encourage the target model to maintain the same local neighborhood structure,
\begin{equation}
    \ell_{\mathrm{CNA}}(x_i) = -\log\frac{\exp[f_T(x_i)\cdot f_T(\aleph(x_i)) / \tau]}{\sum_{j\ne i}^M \exp(f_T(x_i)\cdot f_T(x_j) / \tau)}, \label{eq:loss_cna_u}
\end{equation}
where $\tau$ is the temperature, and $f_T(\cdot)$ the target feature extractor and its output is assumed normalized. Instead of directly minimizing the distance between neighbors, in \cref{eq:loss_cna_u}, the neighbor pairs defined by the source model are pulled together but the non-neighbor pairs will be pushed away during the target model training. In this way, the local structure is preserved, and the knowledge encoded in the structure is transferred to the new model. 

The contrastive loss has the same softmax form as the cross-entropy loss, and this helps when both losses need to be jointly optimized. More importantly, there is no longer a direct requirement to match the source and target outputs. This relaxation enables the CNA to be applicable even when the target model is trained with different loss functions or has different feature dimensions.

While the CNA loss of \cref{eq:loss_cna_u} only considers the nearest neighbor, it can be easily generalized to top-$K$ nearest neighbors,
\begin{equation}
    \ell_{\mathrm{CNA}}^{K}(x_i) =  \frac1{K} \sum_{\tilde{x} \in \aleph^K(x_i)} -\log\frac{\exp[f_T(x_i)\cdot f_T(\tilde{x}) / \tau]}{\sum_{j\ne i}^M \exp(f_T(x_i)\cdot f_T(x_j) / \tau)}, \label{eq:loss_multi_neighbor}
\end{equation}
where $\aleph^K(x_i)$ is the top-$K$ nearest neighbors of sample $x_i$. This formulation simply averages the CNA losses of multiple neighbors. 

\noindent \underline{\emph{Transductive vs. Inductive:}} Traditional manifold learning methods are transductive as they infer the target representation for each data point directly by optimizing the objective. It is advantageous in capturing the manifold of training data since the projection is not limited by any function. But they are hard to generalize to novel data as it requires modification of the original graph. Our \CNA{} is fundamentally different as we learn an inductive target model by making the new model mimic the local structure of the source representation space. In experiments, we show that \CNA{} can also be used to learn an inductive model, and it is non-trivial to adapt the transductive method into an inductive model.

\subsection{Model Distillation/Update}
By distilling the knowledge of the source (teacher) model to the target (student) model, the target model can achieve higher accuracy \cite{hinton2015distilling} or maintain consistency when the model is updated \cite{yan2020positive}. There are various definitions of knowledge which leads to different approaches to transfer it. The objective can be to minimize the sample-wise feature distance between two models with respect to a feature distance function, usually $\ell_2$-distance~\cite{sau2016deep}. Recent works also proposed to use the contrastive loss~\cite{khosla2020supervised} instead of minimizing absolute distance. Our \CNA{} can also be used in these tasks, by carrying the knowledge of the teacher model to the student model by preserving the local structure of the feature space.

Let us consider a classification problem. Let $y_i \in \{1,2, \ldots C\}$ denote the class label for each data sample $x_i$. $h(\cdot,\cdot)$ is the classifier following the feature extractor $f(\cdot)$ to predict the class label, $\hat y = \arg\max_y h(f(x), y)$. In order to learn the full target classification model $\{f_T, h_T\}$ capable of preserving the local topology of the teacher model, the final objective is defined as the weighted sum of the standard cross-entropy loss and the CNA loss:
\begin{equation}
    \mathcal{L}_{\mathrm{CNA}} = \frac{1}{M}\sum_{i}^M\left[\ell_{CE}(x_i, y_i) + \lambda\ell_{\mathrm{CNA}}(x_i) \right] \label{eq:loss_total_cna},
\end{equation}
where $\lambda$ is the trade-off coefficient, and $\ell_{CE}$ is the cross-entropy loss,
\begin{equation}
    \ell_{CE} (x_i, y_i) = -\log \frac{\exp(h_T(f(x_i), y_i))}{\sum_c \exp(h_T(f(x_i), y_c))},
    \label{eq:CE}
\end{equation}
where $y_c = 1,2, \ldots, C$ refers to each class in the datasets. Note that the prediction of the source model $h_S(\cdot, \cdot)$ is not present in the loss function \cref{eq:loss_total_cna}, unlike the distillation term in knowledge distillation methods \cite{hinton2015distilling}. Thus, the two terms in \cref{eq:loss_total_cna} stay congruous even if the source model makes mistakes. As stated before, another major advantage of our proposed \CNA{} loss is the formulation consistency between the two terms, and balancing the two terms in the loss function is relatively easy.

\section{Experiments}
\label{sec:experiments}

We first evaluate the performance of the proposed \CNA{} method on the unsupervised dimensionality reduction task and compare to other manifold learning techniques. \CNA{} is then evaluated on two real-world application tasks: model distillation (from a large to small model) and model update (from a small to a large model).

\subsection{Manifold Learning/Dimensionality Reduction}
In a dimensionality reduction task, the goal is to learn a low-dimensional embedding that captures the manifold in the high-dim data space and favors downstream classifications. The source model $f_S(\cdot)$ reduces to an identity function and the target model $f_T(\cdot)$ is a projector that maps data to a lower dimensional space. We perform the experiments on synthetic datasets and natural datasets. On synthetic datasets, we show qualitative results to demonstrate that \CNA{} is capable of discovering the manifold in a low-dimensional space. On natural datasets, we report quantitative results to measure the quality of the learned embeddings. In this task, \CNA{} is employed in an unsupervised fashion where the cross-entropy loss in \cref{eq:loss_total_cna} is absent.

\noindent {\bf Baselines:} We compare to three transductive manifold learning methods: Isomap \cite{tenenbaum2000global}, LLE \cite{roweis2000nonlinear} and Hessian LLE \cite{donoho2003hessian}. 
We use \CNA{} to learn an inductive function through neural networks on both synthetic and real-world data. We also propose an inductive baseline trained by a modified MVU \cite{weinberger2004learning} objective. Specifically, this baseline uses the same inductive network as \CNA{} and is trained by:
\begin{multline}
    \ell_{MVU} = \frac{1}{M} \sum_{i}^M \Big\{ \frac{1}{K} \sum_{j:x_j \in \aleph^K(x_i)}^M [ d(f_S(x_i), f_S(x_j)) \\ -  d(f_T(x_i),f_T(x_j)) ]^2 -  \gamma \sum_{\substack{j: j\ne i \\ x_j \notin \aleph^K(x_i)}} d(f_T(x_i), f_T(x_j)) \Big\} \label{eq:mvu}
\end{multline}
where $\gamma$ is a balance weight. Basically, this loss function will push non-neighbor data points far away while keeping the distance between neighbors in the target space. During the experiments, we found a small $\gamma$ is necessary (1e-6 e.g.) to avoid trivial solutions where every sample is isolated.

\noindent{\bf Implementations:} For Isomap, LLE and Hessian LLE, the only free parameter is the number of neighbors $K$. We set it to 10 on synthetic data and sweep over [5, 10, 50, 100, 200] on real-world datasets. For MVU and \CNA{}, we use a 3-layer multi-layer perceptron (MLP) with Tanh activation function as the inductive projection model. All networks are trained by Adam optimizer\cite{kingma2014adam} with batch size 256. We fine-tuned the learning rate, number of neighbors $K$, temperature $\tau$ and $\gamma$ in \cref{eq:mvu} on a held-out validation set.

\subsubsection{Synthetic Data}

\begin{table}[tbp!]
    \small
    \centering
    \caption{\small Quantitative results in \% on real-world data. Local-err: local error on training set; 5-NN \Acc: generalization accuracy on test set using a 5-NN classifier trained on the low-dimensional space. The inputs are projected to 40-dim space. *:Hessian LLE uses a 10-dim space due to the constraint on available neighbors \cite{weinberger2004learning}. $\dagger$: \CNA{} can reduce training error but with lower test accuracy.}
    \label{tab:real_data_acc}
    \vspace{-2mm}
    \renewcommand{\arraystretch}{1.1}
    \setlength{\tabcolsep}{1.0mm}
    \scalebox{0.85}{
    \begin{tabular}{l|c|c c|c c}
    \toprule
    \multirow{2}{*}{Methods} & \multirow{2}{*}{Tr./Ind.} & \multicolumn{2}{c|}{MNIST} & \multicolumn{2}{c}{CIFAR10} \\
    & & Local-err & 5-NN \Acc & Local-err & 5-NN \Acc \\
    \midrule
    Isomap & Tr. & 9.3 & 91.2 & 4.2 & 88.7\\
    LLE & Tr. & 10.0 & 91.3 & 3.4 & 88.9 \\
    Hessian* & Tr. & 15.4 & 88.2 & 4.1 & 88.9 \\
    \midrule
    MVU & Ind. & 34.1 & 72.3 & 3.9 & 88.2 \\
    \midrule
    \CNA{}& Ind. (ours) & \bf 8.7 & \bf 92.3 & 3.9 & \bf 89.3\\
    \CNA{}$\dagger$ & Ind. (ours)& - & - & \bf 3.2 & 88.7 \\
    \bottomrule
    \end{tabular}
    }
    \vspace{-4mm}
\end{table}

We performed experiments on three synthetic datasets: S-Curve, Swiss Roll, and Sphere. Data points lie on a two-dimensional manifold in the original 3D space. The plots of the datasets are shown in the left column of \cref{fig:manifold}. All datasets contain 2000 samples. The learning is considered good if the manifold is unfolded and the local structure is retained in the projected 2D space. In the experiments, the MVU and \CNA{} networks have an intermediate layer with 5 nodes and hence 60 total parameters. All networks are trained for 2000 epochs. We visualize the learned manifolds for all the methods in \cref{fig:manifold}.

The results show that the proposed \CNA{} is capable of unfolding the manifold 
on all three synthetic datasets. The data points with similar colors stay neighbors in the projected space. Hence the local topology is well maintained. Transductive manifold learning methods can also unfold the manifold. However, on Swiss Roll dataset, the inductive MVU method fails to retrain the manifold even though we extensively fine-tuned the parameters. This indicates the non-triviality of achieving manifold learning using inductive networks. Notice that although the reconstructed manifolds of \CNA{} for Swill Roll have a non-linear warping, they are not considered poor as the local structure of the two manifolds is identical to that of Isomap and Hessian LLE. Morever, compared to trasductive methods, \CNA{} can learn the manifold using an inductive function with a handful parameters. 

\subsubsection{Real-world Data}
We conducted experiments on two real-world datasets: MNIST \cite{lecun1998gradient} and CIFAR10 \cite{Krizhevsky09learningmultiple}. On MNIST we use the original data (784-dim) as source space, and on CIFAR10 we use the perceptual features (512-dim) of a pretrained ResNet18 model as the source data. For computational efficiency, we randomly select 4000 samples as training data and 1000 samples as test data. The intermediate layer of the MVU and \CNA{} networks has 512 nodes. The networks are trained by 1000 epochs. The data is projected to a 40-dim space except for Hessian LLE because the number of neighbors required is proportional to the square of dimension \cite{weinberger2004learning}. The baseline transductive methods are more competitive, not less, as we found dim=40 works the best for most transuctive methods.

\noindent {\bf Metrics:} We evaluated two metrics: local error and 5-NN generalization accuracy \cite{van2009dimensionality}. Local error is the rate of samples in the training set of which the nearest neighbor is not from the same class. It measures whether the neighbors are maintained in the low-dimensional space. 5-NN generalization accuracy is defined as the accuracy on the test set using a 5-neareast neighbor classifier trained on the projected space. It reflects the quality of the learned data representation regarding the downstream classification task.

The quantitative results on real-word datasets are tabulated in \cref{tab:real_data_acc}. The proposed \CNA{} method shows competitive results across the board. It outperforms all other methods on MNIST and achieves the best 5-NN accuracy on CIFAR10. \CNA{} has higher training local error on CIFAR10. We argue that \CNA{} focuses on broader local topology rather than the nearest neighbor, in order to achieve better generalization when the data is more complex. It can reach lower training error with a sacrifice on the test accuracy (see \CNA$\dagger$ in \cref{tab:real_data_acc}). 
Based on these observations, we conclude that \CNA{} can effectively conduct manifold learning with an inductive function.

\subsection{Real-World Applications}
We evaluated the proposed CNA method on two real-world tasks: model distillation and model update. In both tasks, information from one model is expected to transfer to another model, and our \CNA{} method can be applied by considering the teacher (old) model as source and the student (new) model as target in model distillation (update). ImageNet \cite{DBLP:journals/ijcv/RussakovskyDSKS15} 
and CIFAR100 \cite{Krizhevsky09learningmultiple} are the two major datasets used for evaluation. The used network architectures are the standard ResNet variants \cite{he2016deep} on ImageNet, and ResNet \cite{he2016deep} and ShuffleNet \cite{zhang2018shufflenet,ma2018shufflenet} on CIFAR100 following \cite{tian2020contrastiverep}. 

\noindent{\bf Training Details:} For CIFAR100 (ImageNet), the base initial learning rate was 0.2 (0.1) for batch size of 256, linear scaling \cite{goyal2017accurate} was used for other batch sizes, and cosine learning rate decay schedule was adopted. Weight decay was set as 0.0001 for all ImageNet experiments, but it is sensitive and we searched the best one from $\{0.0001, 0.0002, 0.0005\}$ for CIFAR100. SGD was used for optimization in all experiments. For \CNA{}, we set $\lambda=1.0$ in \cref{eq:loss_total_cna} and temperature $\tau=0.01$ in \cref{eq:loss_cna_u} for all experiments. Each experiment of CIFAR100 (ImageNet) is conducted on 4 (8) V100 GPUs for about 1 (15) hour(s). The results were averaged on 5 runs for all CIFAR100 experiments. More details will be introduced in the following section.

\begin{table}[t]
    \centering
    \caption{\small The model distillation results (top-1 accuracy in \%) on CIFAR100. * denoted results run by us. The other results of the competitors are from \cite{tian2020contrastiverep}.}
    \label{tab:cifar100 distill}
    \vspace{-2mm}
    \renewcommand{\arraystretch}{1.1}
    \setlength{\tabcolsep}{1.2mm}
    \scalebox{0.6}{
    \begin{tabular}{l|c c c c c c}
    \toprule
        Teacher (Source) &ResNet56 &ResNet110 &ResNet110 &ResNet32x4 &ResNet32x4 &ResNet32x4\\
        Student (Target) &ResNet20 &ResNet20 &ResNet32 &ResNet8x4 &ShuffleNetV1 &ShuffleNetV2\\
    \midrule
        Teacher (Source) &72.90 &74.25 &74.25 &79.67 &79.67 &79.67\\
        Student (Target) &69.96 &69.96 &71.26 &72.68 &70.98 &71.76\\
    \midrule
        KD* \cite{hinton2015distilling} &71.82 &71.33 &\bf{73.81} &73.55 &75.05 &75.54\\
        SP \cite{tung2019similarity} &69.67 &70.04 &72.69 &72.94 &73.48 &74.56 \\
        CC \cite{peng2019correlation} &69.63 &69.48 &71.48 &72.97 &71.14 &71.29 \\
        VID \cite{ahn2019variational} &70.38 &70.16 &72.61 &73.09 &73.38 &73.40 \\
        RKD \cite{park2019relational} &69.61 &69.25 &71.82 &71.90 &72.28 &73.21 \\
        LFA* & 71.85 & \bf{71.75} &73.79 &72.41 &75.07 &75.43\\
    \midrule
        CNA (ours) &\bf{71.96} &71.30 &73.63 &74.66 &\bf{75.20} &75.44\\
    \bottomrule
    \end{tabular}
    }
\end{table}

\subsubsection{Baselines}
\label{sec:baseline}

\begin{table}[]
    \centering
    \caption{\small The model distillation results (top-1/top-5 accuracy in \%) on ImageNet.}
    \label{tab:imagenet distill}
    \vspace{-2mm}
    \renewcommand{\arraystretch}{1.1}
    \setlength{\tabcolsep}{0.4mm}
    \scalebox{0.85}{
    \begin{tabular}{c c|c c c c|c}
    \toprule
        &ResNet34 &ResNet18 &KD \cite{hinton2015distilling} &CRD \cite{tian2020contrastiverep} &LFA &CNA (ours)\\
    \midrule
        top-1 &73.31 &69.76 &71.33 &71.17 &\bf{71.41} &71.38\\
        top-5 &91.42 &89.08 &90.35 &90.13 &\bf{90.36} &90.19\\
    \midrule
    \midrule
        &ResNet50 &ResNet18 &KD \cite{hinton2015distilling} &CRD \cite{tian2020contrastiverep} &LFA &CNA\\
    \midrule
        top-1 &76.13 &69.76 &71.25 &- &70.80 &\bf{71.43}\\
        top-5 &93.55 &89.08 &\bf{90.53} &- &90.13 &90.27\\
    \bottomrule
    \end{tabular}
    }
    \vspace{-4mm}
\end{table}

\noindent{\bf Knowledge Distillation} was proposed in \cite{hinton2015distilling} to distill the knowledge from a pre-trained source (teacher) model to a newly trained target (student) model. It forces the target output probabilities to mimic those of the source, by minimizing a KL divergence loss between them,
\begin{equation}
    \ell_{KD} (h_T(x_i), h_S(x_i)) = \tau^2 KL(h_T(x_i)/\tau, h_S(x_i)/\tau),
    \label{eq:loss_KD}
\end{equation}
where $h(\cdot)$ is the classifier prediction, $KL$ is KL divergence and $\tau$ is the temperature.

\noindent{\bf Local Feature Alignment:} To address the issue that LLE \cite{roweis2000nonlinear} lacks the ability of generalizing to points unknown to training data,  \cite{hadsell2006dimensionality} proposed to learn a non-linear mapping to preserve the local neighborhood by contrastive learning. This idea can also be applied here to align the local feature representations, called {\it{local feature alignment}} (LFA). Given the feature representations of the source model $f_S(x_i)$ and the target model $f_T(x_j)$, the idea is to pull them together if $x_i$ and $x_j$ are the same sample, otherwise repel them. The loss is defined as:
\begin{multline}
    \ell_{FA}(x_i, x_j)  = \mathds{1}[i=j] \frac{1}{2} ||f_T(x_j)-f_S(x_i)||_2^2 \\
    + \mathds{1}[i\ne j]\frac{1}{2}[\max(0, \xi-||f_T(x_j)-f_S(x_i)||_2)]^2
    \label{eq:loss_FA}
\end{multline}
where $\mathds{1}[\cdot]$ is an indicator function and $\xi$ is a margin hyper-parameter in the hinge loss term. Similar to \CNA{} loss of \cref{eq:loss_cna_u}, local structure can be aligned by LFA, but the difference is LFA directly optimizes on the $\ell_2$ feature distance.

Similar to \cref{eq:loss_total_cna}, the KD loss of \cref{eq:loss_KD} and FA loss of \cref{eq:loss_FA} is jointly optimized with a cross-entropy loss with trade-off coefficient $\lambda$ in the model distillation task.

\begin{figure}
    \centering
    \includegraphics[width=\linewidth]{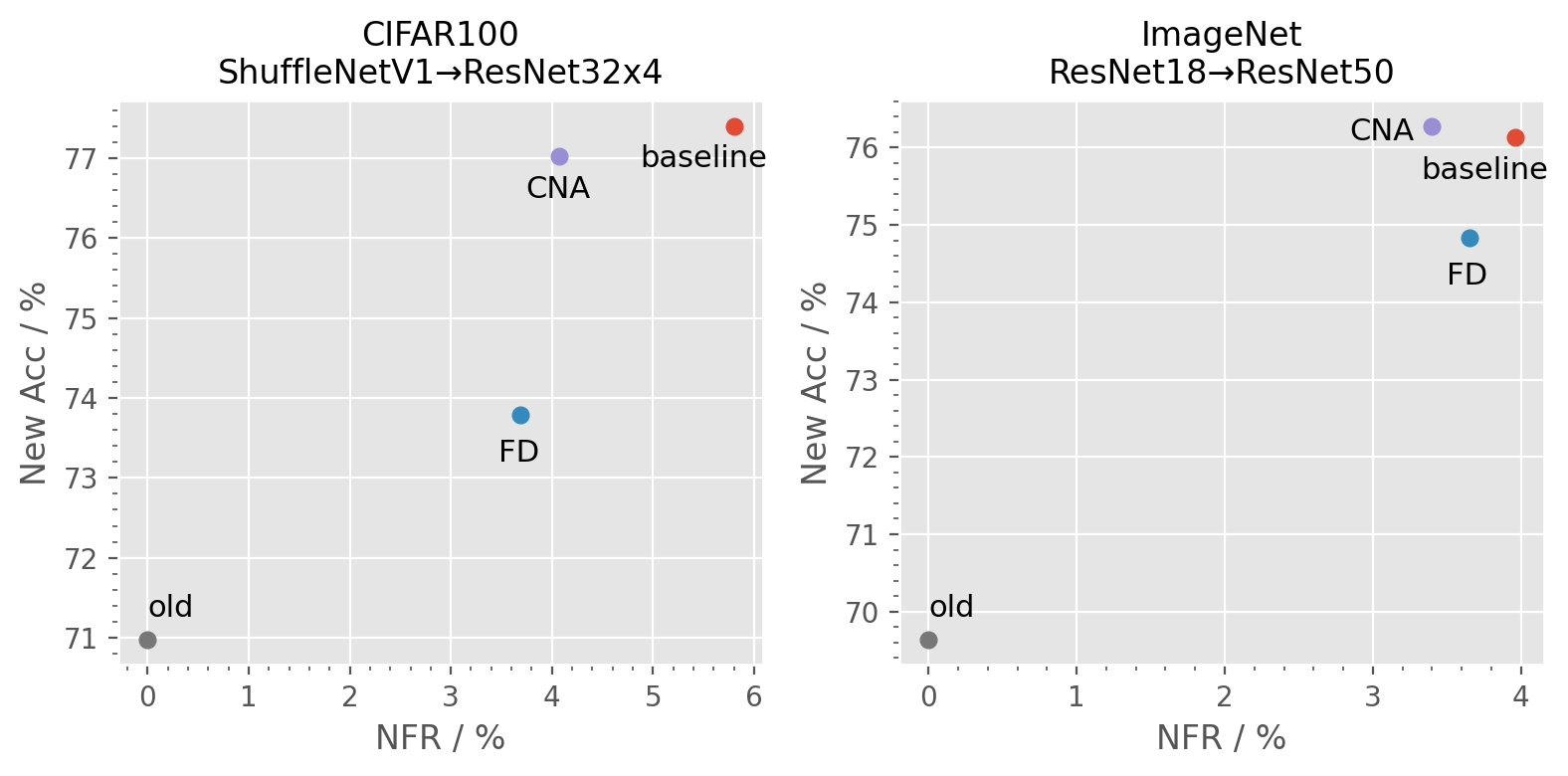}
    \vspace{-5mm}
    \caption{Results of model upgrade on CIFAR100 and ImageNet. X-axis: negative flips rate (NFR) in \%, lower is better. Y-axis: new model accuracy in \% higher is better. The idea model is expected to be located in the top-left corner.}
    \label{fig:diff_model}
    \vspace{-5mm}
\end{figure}

\subsubsection{Model distillation}

We compare CNA to the KD and LFA baselines as introduced in \cref{sec:baseline}, and the recent contrastive representation distillation (CRD) \cite{tian2020contrastiverep}. We followed \cite{tian2020contrastiverep} for the experimental settings for model distillation. In detail, the models on CIFAR100 (ImageNet) were trained for 240 (100) epochs. Differently, we used cosine learning rate scheduling instead of a hand designed schedule. The default batch size is 1024 (512) for CNA and 512 for KD and LFA, on ImageNet (CIFAR100). We re-implemented the KD baseline following \cite{hinton2015distilling}, but set loss trade-off $\lambda=1$ and temperature $\tau=100$, which yielded better results. For LFA, the margin $\xi=3.0$ for all experiments, but we searched for the best trade-off $\lambda$ since it is sensitive. 

The results are shown in \cref{tab:cifar100 distill} for CIFAR100 and \cref{tab:imagenet distill} for ImageNet. In the first part of \cref{tab:cifar100 distill}, the target and source models are of the same architectural
style, but of different styles in the second part. Besides KD \cite{hinton2015distilling}, LFA and CRD \cite{tian2020contrastiverep}, some recent works of SP \cite{tung2019similarity}, CC \cite{peng2019correlation}, VID \cite{ahn2019variational} and RKD \cite{park2019relational} are also compared. First, we found that our KD and LFA baselines already achieved very good results, better than SP, CC, VID and RKD in most of cases, and comparable to CRD. But the results of LFA are sometimes unreliable. For example, when distilling ``ResNet32x4'' to ``ResNet8x4'' on CIFAR100 or distilling ``ResNet50'' to ``ResNet18'' on ImageNet, LFA underperforms KD, CRD and CNA by a large margin. This is probably due to the direct optimization of LFA on the $\ell_2$ feature distance. Next, our \CNA{} achieved reliable results comparable to KD and CRD, although CNA only tries to preserve the local topology, irrespective of the representation knowledge as distilled by KD. This shows the effectiveness of the proposed CNA even when it is evaluated for a classification task. The observations are similar in the experiments of ImageNet in \cref{tab:imagenet distill}, showing that the CNA can generalize well on different datasets.

\subsubsection{Reducing Regression in Model Upgrade}
In~\cite{yan2020positive} the problem of regression in model upgrade is introduced. A variant of KL-divergence based distillation loss was used to reduce the amount of new errors introduced by a new model in comparison to an old reference model. 
We expect the capacity of CNA in preserving the structure knowledge 
would also help reduce the regression. 
We compare CNA to (i) a naive baseline approach to train the new model using the standard cross-entropy loss -  no attempt to address model regression is considered in this case (ii) the latest approach using focal distillation (FD) \cite{yan2020positive}.

\begin{table}[t]
    \centering
    \caption{Ablation studies on batch size.}
    \label{table:batch size}
    \vspace{-2mm}
    \renewcommand{\arraystretch}{1.1}
    \setlength{\tabcolsep}{1mm}
    \scalebox{0.85}{
    \begin{tabular}{l c|c c c c c}
    \toprule
        batch size &32 &64 &128 &256 &512 &1024\\
    \midrule
        CIFAR100 &71.51 &71.36 &71.87 &71.79 &\bf{71.96} &-\\
        ImageNet &- &- &70.75 &70.87 &71.08 &\bf{71.38}\\
    \bottomrule
    \end{tabular}
    }
\vspace{-3mm}
\end{table}

\begin{table}[]
    \centering
    \caption{Ablation studies on Number of MLP.}
    \label{table:mlp}
    \vspace{-2mm}
    \renewcommand{\arraystretch}{1.1}
    \setlength{\tabcolsep}{2mm}
    \scalebox{0.85}{
    \begin{tabular}{l c|c c}
    \toprule
        \# MLP &0 &1 &2\\
    \midrule
        CIFAR100 &\bf{71.96} &71.56 &71.49\\
        ImageNet &\bf{71.38} &71.01 &71.00\\
    \bottomrule
    \end{tabular}
    }
    \vspace{-3mm}
\end{table}

\noindent{\bf Metrics:}
We evaluated the negative flip rate (NFR) \cite{yan2020positive} and the accuracy for each model. The NFR is defined as the fraction of negative flips:
\begin{equation}
    \mathrm{NFR} = \frac1{N} \sum_{i=1}^{N} \mathds{1}[\hat y_n^{(i)} \ne y_i \; and \;\; \hat y_o^{(i)} = y_i]
\end{equation}
where $\hat y_n^{(i)}$ and $\hat y_o^{(i)}$ are the predicted labels of the new model and the old model for input image $x_i$, and $N$ is the number of samples in the test set $\mathcal{S}_{test}=\{(x_i,y_i),i=1..N\}$.

All models in these experiments were trained with batch size 512. On CIFAR100 (ImageNet), the model was trained for 240 (90) epochs, with learning rate decreased by 0.1 every 80 (30) epochs. For CNA, the trade-off $\lambda$ in \cref{eq:loss_total_cna} is chosen by validation as $1$ for CIFAR100 and $0.2$ for ImageNet. To ensure a fair comparison, the hyper-parameters for focal distillation were also chosen through the same validation split of CNA.

We consider the common model update scenario: the new model architecture is
larger than that of the old model. We evaluated a variety of architecture changes, including ShuffleNet and ResNet variants.
The models were both trained on the same dataset. The results are visualized in \cref{fig:diff_model}.
We observe that CNA can consistently reduce the NFR. In particular, the relative NFR reduction compared to the baseline new model is 29.9\% on CIFAR100 and 14.1\% on ImageNet.
On ImageNet, CNA slightly improves the model accuracy from 76.13\% to 76.28\%.
CNA outperforms focal distillation by obtaining higher accuracy with comparable or lower NFR.
This performance improvement can be attributed to the benefits of imposing direct representation regularization instead of mimicking the input-output behavior via knowledge distillation. The latter in the scenario of model upgrade would cause inevitable accuracy degradation of the new model, which the CNA does not suffer from. 

\subsection{Ablation Studies}
\label{sec:ablation}

\begin{table}[b]
\vspace{-2mm}
    \centering
    \caption{\small Ablation studies on temperature.}
    \label{table:temperature}
    \vspace{-2mm}
    \scalebox{0.85}{
    \begin{tabular}{l c|c c c}
    \toprule
        temperature &0.001 &0.01 &0.05 &0.1\\
    \midrule
        CIFAR100 &71.42 &\bf{71.96} &71.44 &70.40\\
        ImageNet &70.87 &\bf{71.38} &70.84 &70.52\\
    \bottomrule
    \end{tabular}
    }
\end{table}

\begin{table}[b]
    \centering
    \caption{\small Ablation studies on number of neighbors $K$.}
    \label{table:neighbor}
    \vspace{-2mm}
    \scalebox{0.85}{
    \begin{tabular}{l c |c c c}
    \toprule
        $K$ &1 &2 &4 &8 \\
    \midrule
        CIFAR100 &\bf{71.96} &71.94 &71.50 &71.41\\
        ImageNet &\bf{71.38} &71.26 &71.14 &71.02\\
    \bottomrule
    \end{tabular}
    }
\end{table}

To further understand the performance of CNA, we conducted a series of model distillation ablation studies. The source/target model is ResNet56/ResNet20 on CIFAR100 and ResNet34/ResNet18 on ImageNet. The default settings are: temperature $\tau=0.01$, number of neighbors $K=1$, number of MLP is 0 and batch size is 512 (1024) for CIFAR100 (ImageNet).

\noindent{\bf Batch Size.} The effect of batch size is investigated in \cref{table:batch size}. It can be found that the CNA algorithm is quite robust to different batch sizes on CIFAR100, as long as the batch size is not smaller than 128. But the performance decreases when the batch size becomes smaller on ImageNet. This is probably because there are 1,000 classes for ImageNet and it is possible there is no very close neighbor in a single batch if the batch size is too small. Relatively large batch size is preferred for CNA to better preserve the local structure.

\noindent{\bf Temperature $\tau$} was ablated in \cref{table:temperature}. It shows the best choice is around $\tau=0.01$. When it is decreased or increased too much, the performances drop. This temperature is also robust to different datasets.

\noindent{\bf Number of neighbors $K$} in \cref{eq:loss_multi_neighbor} was ablated in \cref{table:neighbor}. On both datasets, $K=1$
performs considerably well, but the performances decrease when more neighbors were counted in the loss function of \cref{eq:loss_multi_neighbor}. This is probably
because the contrastive loss formulation (\cref{eq:loss_multi_neighbor}) is curated for single pair contrast. Directly including the other neighbors without considering their further structure information, e.g. neighbor ranking, from the old model may impose inaccurate neighborhood regulation on the new model, thus degrading the performance.

\noindent{\bf Number of MLPs.} The hidden MLP layers have been shown to be very helpful for representation learning \cite{chen2020simple,he2020momentum}. Results from investigations of its effect on CNA are shown in \cref{table:mlp}. It shows that the hidden MLP layer does not help in CNA. The possible reason could be that the goal of CNA, maintaining the local data structure, is different from the representation learning \cite{chen2020simple,he2020momentum}.

\section{Conclusion}
\label{sec:conclusion}
In this paper, we proposed a new method, contrastive neighborhood alignment (\CNA), that preserves the local structure of feature spaces between models. The effectiveness of \CNA{} is illustrated on three problems: manifold learning, where the topology of the original data is maintained in a low-dimensional space using an inductive model, model distillation, where a compact student is trained to mimic a larger one, and model upgrade, where the new model is more powerful than the old one. 
Existing methods in model distillation are primarily focused on matching the output predictions/logits for the pair of models in question whereas \CNA{} is motivated differently by preserving the local neighborhood structures of the samples in the representation spaces. The instance-level contrastive loss in \CNA{} is empirically shown to work harmoniously with the standard cross-entropy loss.

\noindent {\bf Limitations:} The concept of aligning the local neighborhoods is intuitive but its performance gain in model distillation still has room to improve. Nonetheless, \CNA{} provides a new means for preserving the integrity of the model by matching the feature manifold using a contrastive term that has not been previously explored. 

\noindent {\bf Potential Negative Societal Impacts:} \CNA{} may accidentally transfer the bias in the source feature embedding \cite{bolukbasi2016man} to the target model. As a result it may unintentionally amplify these biases when deploying the model. Selecting unbiased source model or applying debiasing algorithm to the learned new model can mitigate this negative effect.

{\small
\bibliographystyle{ieee_fullname}
\bibliography{egbib}
}

\end{document}